\documentclass{IEEEtaes}
\IEEEoverridecommandlockouts
\usepackage{graphicx}
\usepackage{amsmath}
\usepackage{algorithm}
\usepackage{algpseudocode}
\usepackage{units}
\usepackage{booktabs}
\usepackage{url}
\usepackage{caption}
\usepackage{subcaption}
\usepackage{float}
\usepackage{capt-of}
\usepackage{multicol}

\usepackage[absolute]{textpos}
\usepackage[normalem]{ulem}

\jvol{62}
\pubyear{2026}
\doiinfo{TAES.2026.3683620}

\makeatletter
\def\taesjournalline{{\rffont IEEE TRANSACTIONS ON AEROSPACE AND ELECTRONIC SYSTEMS\hspace*{7pt} VOL.\ \@jvol\hspace*{7pt} \@pubyear}}
\def\ps@plain{\def\@oddhead{}\def\@evenhead{}%
  \def\@evenfoot{\hbox to \textwidth{{\rffont\thepage}\hfill\taesjournalline}}%
  \def\@oddfoot{\hbox to \textwidth{\taesjournalline\hfill{\rffont\thepage}}}}
\let\taesorigheadings\ps@headings
\def\ps@headings{\taesorigheadings
  \def\@evenfoot{\hbox to \textwidth{{\rffont\thepage}\hfill\taesjournalline}}}
\ps@headings
\def\leftmark{PHILLIPS ET AL.}\def\rightmark{ROBUST SAFETY ASSURANCE FOR ROBOTIC SWARMS}
\makeatother

\usepackage{eso-pic}
\AddToShipoutPictureBG*{%
  \AtPageUpperLeft{\put(\LenToUnit{-\hoffset},\LenToUnit{\dimexpr\voffset-17pt\relax}){\makebox[\paperwidth]{\parbox{0.92\paperwidth}{\centering\scriptsize
  \copyright~2026 IEEE. Personal use of this material is permitted. Permission from IEEE must be obtained for all other uses, in any current or future media, including reprinting/republishing this material for advertising or promotional purposes, creating new collective works, for resale or redistribution to servers or lists, or reuse of any copyrighted component of this work in other works.}}}}}

\title{Robust Safety Assurance for Robotic Swarms Using Decentralized Collision Verification}

\author{Grant Phillips}
\member{Student Member, IEEE}
\affil{University of Nebraska-Lincoln, Lincoln, NE USA}

\author{Hoang-Dung Tran}
\affil{University of Florida, Gainesville, FL USA}

\author{Justin Bradley}
\member{Senior Member, IEEE}
\affil{North Carolina State University, Raleigh, NC USA}

\receiveddate{Manuscript received 6 March 2025; revised 8 February 2026; accepted 22 March 2026. This is the authors' accepted manuscript; the published version is available at \href{https://doi.org/10.1109/TAES.2026.3683620}{https://doi.org/10.1109/TAES.2026.3683620}.\\
This work was supported by the National Science Foundation under grants NSF-2514761 and NSF-2221648, the National Strategic Research Institute grant NSRI FA4600-20-D-0003, and the United States Department of Agriculture grant USDA 2023-67021-38977.
}

\corresp{
{\itshape (Corresponding author: G. Phillips)}.}

\authoraddress{Grant Phillips is with the School of Computing at the University of Nebraska-Lincoln, Lincoln, NE USA
(e-mail: \href{mailto:grant.phillips@huskers.unl.edu}{grant.phillips@huskers.unl.edu}). Hoang-Dung Tran is with the Department of Computer and Information Science \& Engineering at the University of Florida, Gainesville, FL USA (e-mail: \href{mailto:dungtran@ufl.edu}{dungtran@ufl.edu}). Justin Bradley is with the Computer Science Department at North Carolina State University, Raleigh, NC USA (e-mail: \href{mailto:jmbradl5@ncsu.edu}{jmbradl5@ncsu.edu}).}

\supplementary{Video of swarm experiment: \href{https://youtu.be/Y-Z3IN--_Vc}{Video Demonstration}. Project source code: \href{https://github.com/unl-nimbus-lab/arl-swarm/tree/verifier_2025}{link to github}.}

\begin{document}

\maketitle

\begin{abstract}

In this paper we present the methodologies, results, and analysis used to deploy a decentralized safety verification algorithm on a robotic swarm to prevent collisions at runtime in an outdoor environment.
Typically, approaches toward verification of swarms of cyber-physical systems (CPS) are evaluated solely in numerical simulations or in highly controlled, noiseless indoor environments.
Our efforts to implement collision avoidance on a real swarm system outdoors have revealed that the underlying algorithmic assumptions juxtaposed against the implementation details of the target system are critical factors in a method's efficacy in the real world.
Here we modify, enhance, and analyze a decentralized reachability-based method for collision avoidance on a size, weight, and power-constrained uncrewed aerial vehicle (UAV) swarm with the objective of illustrating the challenges of real-world CPS deployments as well as provide guidance for researchers in this space.
Verification performance, resource utilization, and timing data were gathered over the course of 140 total outdoor vehicle flights, including 6 swarm flights composed of 8 UAVs,
and used to improve collision-avoidance performance as experimentation progressed.
Our extensive experiments show that the proposed decentralized runtime verification approach can efficiently verify the safety of multiple UAVs in real-time and help all agents take safe actions to avoid collisions.
On average, each agent, using our open-source \texttt{safety\_checker} library, takes approximately \unit[500-520]{ms} to verify its safety for the next \unit[4]{s} into the future, in which the reachable set broadcast time dominates the total verification time ($\sim$\unit[300-500]{ms}).

\end{abstract}

\begin{IEEEkeywords}
    Autonomous Aerial Vehicles, Formal Verification, Swarm Robotics
\end{IEEEkeywords}

\section{Introduction}\label{sec:introduction}

\begin{figure}[h]
    \centering
    \includegraphics[scale=0.25]{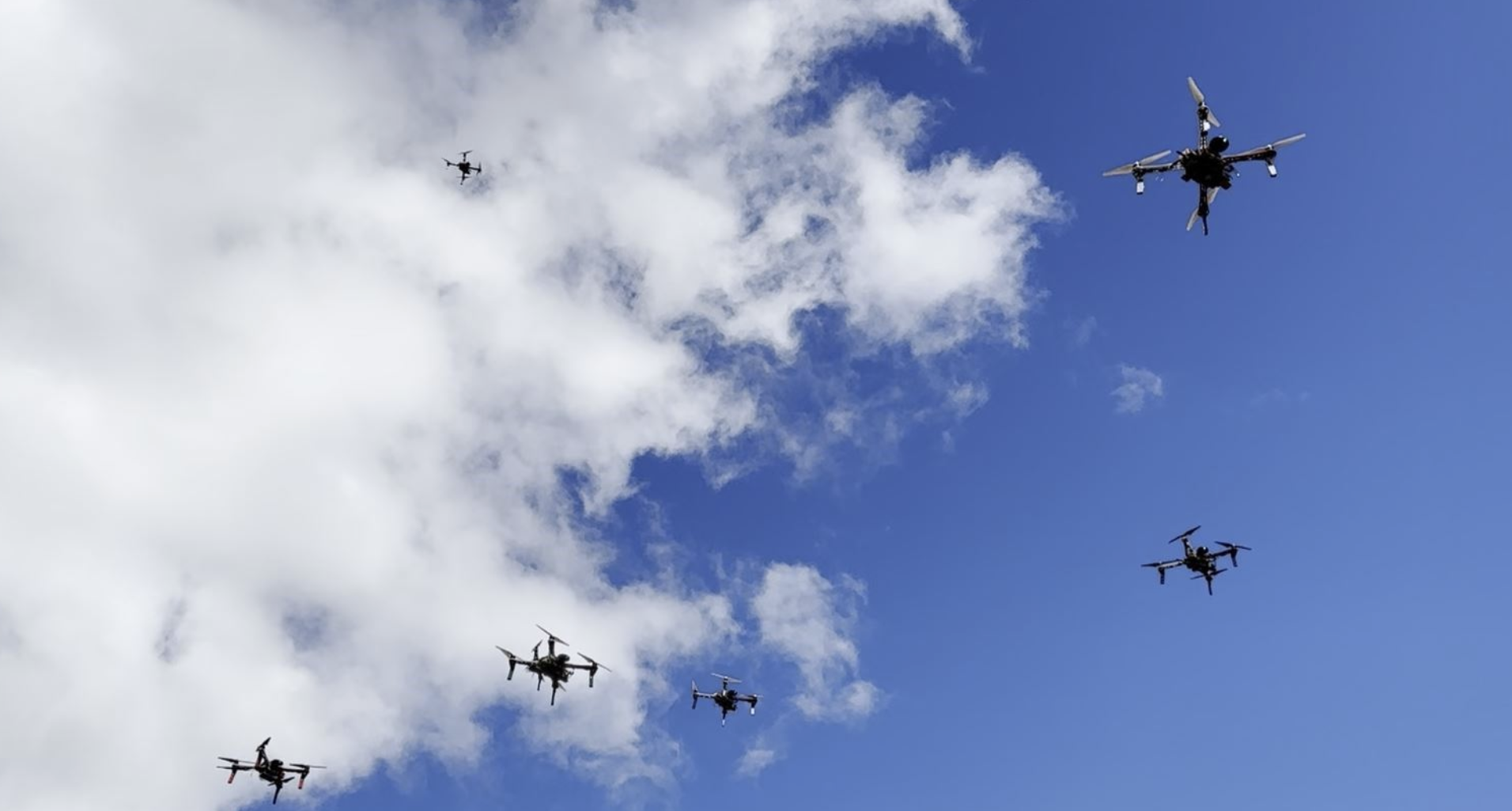}
    \caption{UAV agents as part of a swarm operating in a representative environment.}
    \label{fig:swarm_pic}
\end{figure}

Many of the emergent strategies in the verification community have only been exercised either in numerical simulation or, in the case of robotics, indoor conditions in which localization is accomplished by a high-precision motion capture (MoCap) system, and communication occurs via a local WiFi setup. These conditions enhance the ability to isolate the performance of the verification strategy and algorithm but also remove two of the most significant complicating factors of outdoor, real-world deployments where uncertainty and disturbances in the environment, as well as much less capable communication and localization systems may dominate design decisions and algorithm performance. Furthermore, when it comes to robotic swarms the \textbf{vast majority} of research only validates in either simulation or indoor testbeds, which, we argue, is significantly holding back technological advancement in this space by widening the gap between theory and practice. In a survey of 330 research papers\footnote{Selection chosen by papers tagged with ``multi-robot'' or ``swarm'' in the last 5 years of IEEE IROS, ICCPS, ICRA, and CDC conferences.} presenting novel swarm control, verification, or task management, only 8\% performed any outdoor experimentation, and only 1\% on swarms of more than 3 Unmanned Aircraft Vehicles (UAVs). Simulation results can be insightful, but limitations in vehicle hardware can break design-time assumptions, leaving the usefulness of methods without experimental results in question, particularly those motivated by real-world applications (i.e., search and rescue in disaster areas or contested and dynamic spaces).

Aircraft, to some extent, rely on the large-sky, small-plane paradigm to limit collisions. This breaks down near airports and congested flight corridors, and completely falls apart in swarm \cite{arnold2019what} operations where mission objectives are furthered by close proximity flight. Crewed aircraft rely on see-and-avoid \cite{kephart2008comparison}, Traffic Collision Avoidance Systems (TCAS) \cite{williamson1989developmenta}, and separation by air traffic controllers \cite{brooker1983aircraft} to remain safe, but these solutions are not suited for autonomous, small UAVs due to their Size, Weight, and Power (SWaP)-constrained nature. Moreover, those solutions attempt to maximize separation distance within constraints, while swarms often attempt to fly in close formation to accomplish particular goals.

Formal-method-based verification strategies offer promise in this regard as they provide more precise spatial and timing limits for vehicles. However, such strategies are also traditionally computationally complex, and may require accurate dynamic models of the collective swarm to ensure collision-free trajectories \cite{zhou2015reachable}. Decentralized verification strategies are scarce and, when available, may require highly reliable, high-rate communication, complex models, or abundant computation, all of which are poorly suited to SWaP-constrained UAVs.

The objective in this work is to examine the adaptations needed to implement a reachability-based verification strategy onboard a modern robotic swarm fleet, as well as to provide extensive experimental data and analysis to inform future research into assured autonomy strategies. In this paper we focus narrowly on a swarm anti-collision strategy, though our analysis can be generalized since we capture computational and communication resource limitations. The goal is to implement, integrate, and deploy the verification strategy and obtain data about: 1) what simplifications and optimizations must be made to obtain suitable online, runtime performance on a common small UAV platform, and 2) what unforeseen challenges exist in a real-world deployment, and how the deployment data can be used to further improve the system. To do this, we enhance, and adapt the reachability-based approach described in \cite{tran2019decentralized} for implementation due to its decentralized nature and suitability for a swarm deployment in which the desired global safety property is to prevent potential collisions between the agents in the swarm.

This work bridges the gap between theoretical formal verification and deployment in a representative environment. While the core mathematical reachability analysis builds upon the face-lifting method established in \cite{tran2019decentralized}, our contribution lies in the system-level architectural synthesis and rigorous characterization required to deploy these guarantees on hardware. Specifically, this paper contributes:

\begin{itemize}
    \item The synthesis of a runtime verification framework that adapts formal reachability analysis to the specific timing, network, and SWaP constraints of commercial-off-the-shelf (COTS) UAV hardware.
    \item Extensive outdoor evaluation (140 flights across 6 swarm deployments) demonstrating the efficacy of decentralized verification in environments with realistic disturbances and communication loss, exceeding the scale of prior experimental work (e.g., \cite{gillula2011applications}).
    \item A detailed decomposition of Worst-Case Execution Time (WCET) and network broadcast delays, providing design guidelines for the trade-offs between swarm size, safety horizons, and verification periods
\end{itemize}

\section{Related Work}\label{sec:related-work}

\textbf{Reachability-based verification.}
Reachability analysis is the heart of many verification tools for cyber-physical systems (CPS). It involves the construction of exact or overapproximate reachable sets containing all possible trajectories of the systems.
Many verification tools based on reachability algorithms have been proposed for CPS modeled as hybrid linear/nonlinear systems, such as SpaceEx \cite{frehse2011spaceex}, d/dt \cite{asarin2002d}, JuliaReach \cite{bogomolov2019juliareach}, Hylla \cite{bak2017hylaa}, CORA \cite{althoff2015introduction}, Hypro \cite{schupp2022recent}, Flow* \cite{chen2013flow}, DReach \cite{kong2015dreach}, C2E2 \cite{duggirala2015c2e2}, DryVR \cite{fan2017dryvr}, Hamilton-Jacobi \cite{bansal2017hamilton}.
Distinguished from the reachability-based verification approach, Keymaera \cite{platzer2008keymaera} is the well-known theorem prover for CPS.
Recently, there has been an emerging research area in reachability-based verification for deep neural networks
\cite{tran2019star,yang2021reachability,zhang2020verification},
and learning-enabled CPS \cite{tran2020verification}.

Although there has been rigorous research on CPS verification, only a few approaches focus on verifying distributed CPS \cite{kumar2012hybrid,loos2011adaptive,li2023verse,duggirala2012static}.
Importantly, these methods require \emph{explicit modeling} of distributed CPS as a hybrid system and perform the verification in a \emph{centralized manner}. We seek a simpler, decentralized verification approach as it enhances robustness of the distributed system and is more suitable for field-based drone swarm operations where centralized strategies may be impossible (i.e., due to range). Particularly, we use the reachability-based face-lifting method \cite{dang1998reachability} to compute the reachable set of each autonomous agent to a time horizon \unit[$T$]{s} in the future. Importantly, this method does not require \emph{explicit modeling} of the distributed CPS, and instead we perform distributed reachability analysis via reachable set communication between the agents \cite{tran2019decentralized}. The face-lifting method can be more efficient for real-time verification on small embedded platforms as it does not require complicated data structures/libraries or extensive computational resources \cite{bak2014sfl}.

\textbf{UAV Collision Avoidance.}

In the broader aviation context, the Traffic Alert and Collision Avoidance System (TCAS II) \cite{williamson1989developmenta} and the next-generation Airborne Collision Avoidance System (ACAS Xu) \cite{owen2019acasxu} for uncrewed systems represent the gold standards for mid-air de-confliction. While TCAS II relies on rule-based logic to issue vertical resolution advisories, ACAS Xu utilizes probabilistic models (Markov Decision Processes) to optimize avoidance logic \cite{mueller2016multi}. However, these centralized or pairwise standards are primarily designed for "detect-and-avoid" operations to ensure large separation minima in civil airspace. This design philosophy often proves too conservative for robotic swarms, where mission objectives frequently require close-proximity formation flight that would trigger constant false alarms in separation-based systems. Furthermore, standard ACAS implementations typically rely on pre-computed lookup tables or heavy centralized computation, which may not align with the dynamic, fully decentralized, and SWaP-constrained architecture required for scalable autonomous swarms. In contrast, our reachability-based approach provides formal set-based safety guarantees that allow agents to operate safely within tighter spatial margins. Collision avoidance for UAVs has been extensively studied as the platform has gained popularity over the past two decades. Towards integrating UAVs into the national airspace, the vast majority of the work is focused on the ability of a single UAV to detect and avoid collisions with other air traffic \cite{yasin2020unmanned} or obstacles \cite{gageik2015obstacle}. In this context, perception \cite{zsedrovits2011collision,mcfadyen2016survey} is responsible for detection and identification of the obstacle and some combination of an estimator \cite{luo2013uav}, path planner \cite{lin2017samplingbased,tony2017avoidance}, and controller \cite{herbert2017fastrack} generates and guides the vehicle through safe trajectories.
\begin{figure*}[t]
    \centering
    \includegraphics[width=.95\linewidth]{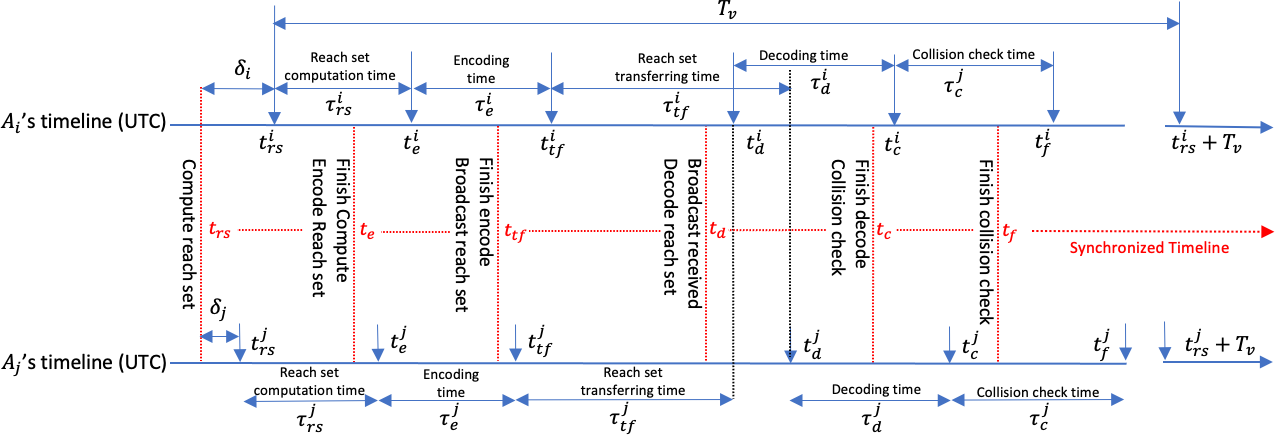}
    \vspace{-6pt}
    \caption{Collision avoidance sub-tasks during a verification period for two agents' local timelines. Note that although $A_i$ and $A_j$ local clocks are synchronized to UTC, local events $t^i_{rs}$ and $t^j_{rs}$ do not necessarily take place at the same global time. Each verification period begins with a reach set computation, and is immediately followed by encoding and broadcasting. The remainder of $T_v$ is used to decode messages and check if any decoded messages indicate a potential collision \cite{tran2019decentralized}.}
    \label{fig:reach_timeline}
\end{figure*}

Aerial swarm collision avoidance has received comparatively less attention, in part due to the need to deliberately fly near friendly agents in the swarm to accomplish mission objectives. This is most typically researched as cooperative formation flying that incorporates collision avoidance into the controller either by a combined controller \cite{sharma2009collision,bai2020hierarchical}, potential fields \cite{abeywickrama2017algorithm}, machine learning \cite{wang2007cooperative}, or model predictive control \cite{Jin2022}. Of particular relevance is the work by Gillula et al. \cite{gillula2011applications}, which applied hybrid reachability analysis to a 4-vehicle swarm (STARMAC). While \cite{gillula2011applications} validated the feasibility of reachability-based safety, our work differs in three critical system-level aspects:
\begin{enumerate}
    \item We implement a fully decentralized architecture where reachable sets are broadcast directly, reducing computational redundancy compared to state-broadcasting approaches.
    \item We validate on a larger 8-vehicle swarm in outdoor conditions subject to GPS noise and wind gusts, whereas previous validations were primarily indoor or limited to fewer agents.
    \item Unlike \cite{tran2019decentralized} which introduced the theoretical decentralized face-lifting protocol, this work provides the necessary empirical analysis of computation and communication overheads that limit practical deployment, offering a blueprint for transitioning formal methods from simulation to the field.
\end{enumerate}

Our work builds on this with a fully decentralized approach wherein the reachable spaces are themselves broadcast thereby leveraging communication to reduce computational workload for each agent.
We demonstrate our approach outdoors using a 8-vehicle swarm avoiding all collisions, as well as performing full timing analysis on communication and computation, and the impact of packet loss on performance.

\section{Decentralized Verification} \label{section:decentralized_verification}

Our deployed collision avoidance scheme is modified from the approach described in \cite{tran2019decentralized}. Early attempts to adapt the proposed algorithm to a real system revealed that real-world factors, most significantly resource constraints, greatly limit what is possible. For instance, even under significant algorithm simplification the code, originally written in python, could not compute and verify reachable sets quickly enough, and that a centralized approach was too unreliable due to limitations of SWaP-constrained radios. These factors influenced our approach and motivated our timing analysis on the data collected from outdoor deployments, particularly since most similar methodologies are not evaluated on a real system. The modified algorithms described in this section have been implemented within the robot operating system (ROS) \cite{Quigley09} framework and deployed on a distributed system of custom-built, quadrotors, or ``agents.''

\subsection{System Overview}
\label{subsec:system_overviewd}

The decentralized  platform used for testing our safety application is made up of 8 identical, custom 450-sized quadrotors. The low-level flight controls are governed by a Cube-Blue flight control unit (FCU) while the collision avoidance application executes on a Raspberry Pi 4 companion computer. Inter-agent communications are provided by a network of XBee radios configured as a mobile ad-hoc network (MANET), allowing the agents to send and receive verification-related data with each other on a first-come-first-serve basis. Global position information is gathered by a here3 GPS module with and \unit[8]{Hz} update rate. For safety purposes, each vehicle is fitted with a second long-range, low-bandwidth radio used exclusively for agent-to-ground control communication.

\subsection{Verification Overview}

Swarm safety is achieved through distributed, periodic verification of each agent's projected states. This approach to safety verification is structured such that all collision detection tasks take place in a \textit{verification period} $T_v$; representative verification periods for two networked swarm agents, agent $A_i$ and $A_j$,  are shown in Figure \ref{fig:reach_timeline}. The safety checking mechanism, known as peer-to-peer collision detection, requires all agents in the swarm to compute and broadcast a reachable set of states each $T_v$. A reachable set of states, or reach set, is a set of all possible states that a system can attain from an initial state over a time-horizon given its current trajectory and a motion model of the agent. In this work, a reachable set is approximated by a hyper-rectangle, $\mathcal{R}$, which is encoded as a set of intervals that bound the ``reachable'' values for each system state variable. The reach set also includes the time, $t$, it was computed and the reach-time horizon, $t+T$, to which it was computed forward; the complete reachable set of states is denoted as $\mathcal{R}[t,t+T]$.

When a reach set is computed, it is encoded and broadcast to the other agents in the swarm (Figure \ref{fig:reach_timeline}, timestamps $t_e$ to $t_d$). When a reach set is received, the reach-time horizon is checked to confirm the set has not ``expired'' due to transmission delays. If the received set is not expired, it is checked for any intersection with the receiving agent's reach set (Figure \ref{fig:reach_timeline}, timestamps $t_d$ to $t_f$). Intersections are flagged as potential collisions and trigger avoidance action.

The anti-collision application's performance is subject to the reachable set's characteristics along with the communication network's reliability and the number of agents in the swarm $N$. Discussion related to the impact of the communication network and the number of agents in the swarm can be found in Section \ref{section:experimental_setup_and_evaluation}. The remainder of this section describes how the reach set is computed and used for decentralized safety verification.

\subsubsection{Reachable Set Computation}

\begin{figure}[h]
    \title{}
    \centering
    \includegraphics[width=0.9\linewidth]{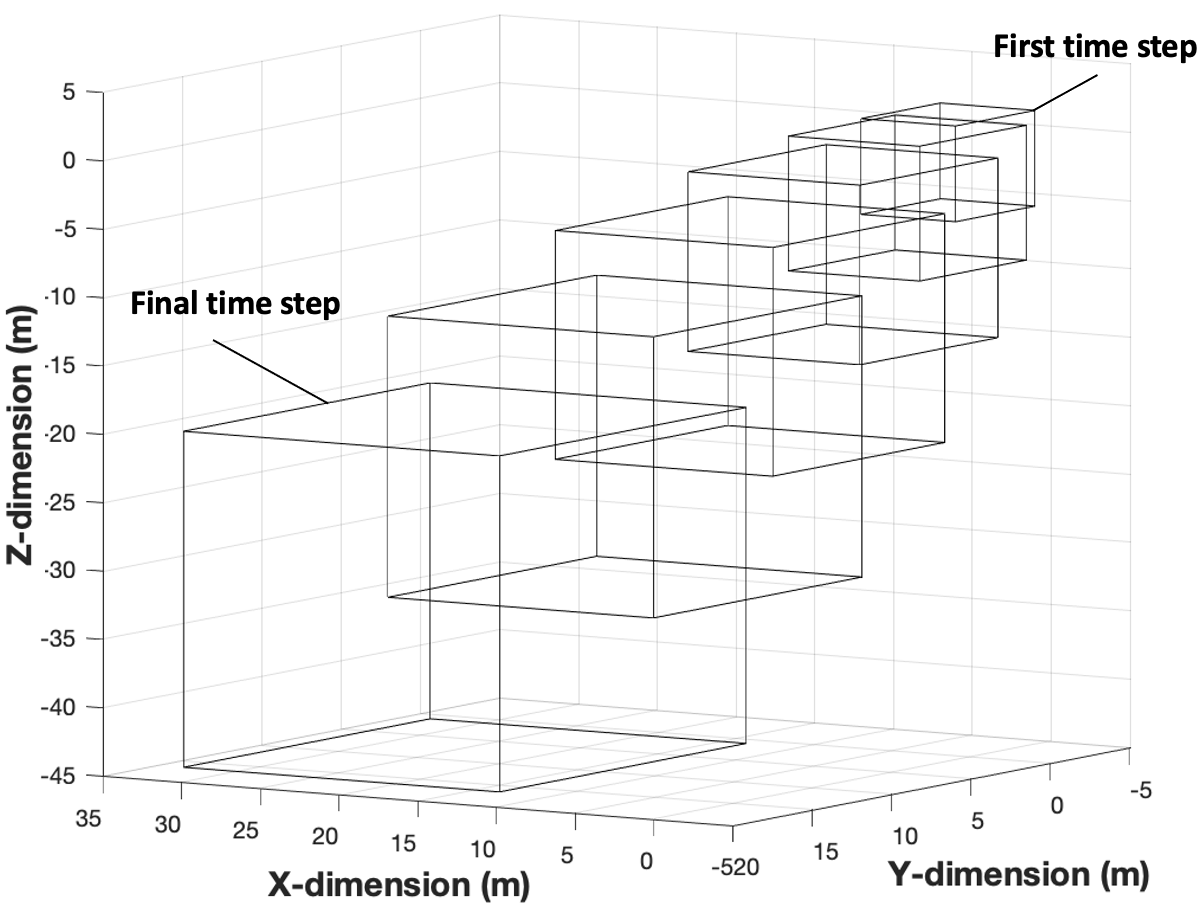}
    \caption{The reachable set over the time period [$t,t+T$] is computed by calling the single face-lifting operation on smaller time steps. The result is a more accurate estimate of the reachable set.}
    \label{fig:in_between_sets}
\end{figure}

The reachable set of states is computed using a process known as the face-lifting method \cite{dang1998reachability,bak2014sfl}. The face-lifting (FL) method computes reachable set over the period [$t,t+T$] by projecting a set of initial conditions, $I$, forward in time given a motion model of the system. In this work, we define the initial set as the agent's current state estimate $x = (x, y, z, v_x, v_y, v_z)$ $\pm$ GPS error. The core step of the face-lifting method -- the single face-lifting (SFL) \textit{operation} -- is used to compute the reach set. The SFL produces an over-approximation of the reachable set, which for safety applications, increases the safety margin, but may lead to unwanted false detections due to excessive over-approximation. We can improve a reachable set's accuracy by dividing the reach-time horizon into smaller time steps of size $h$ and calling the SFL operation on each time step as shown in Figure \ref{fig:in_between_sets}. By using the hull (extreme dimensions) of the resulting reachable sets computed from the smaller time steps, we produce a more accurate estimate of the reach set over the whole time period; the process of dividing the time period into smaller steps is known as the face-lifting method.

The size of the time step $h$ directly affects the run time of the FL method. For large time steps, the resulting reach set is a large over-approximation; however, if the time step is too small, the FL method may run too long and risks overloading the verification period. Reach set accuracy and computation time can be balanced by specifying a maximum run time, $T_{runtime}$, for reach set computation with a large initial time step of size $h_i$. The FL method is run once with time step, $h_i$, and the remaining computation time is evaluated; if any run time remains, the step size is halved and the FL method is called again on the same time period but with the smaller time step. This is repeated until the remaining run time reaches zero or the the step size is reduced to a pre-determined minimum $h_{min}$.

\begin{figure}[h]
    \title{}
    \centering
    \includegraphics[width=0.95\linewidth]{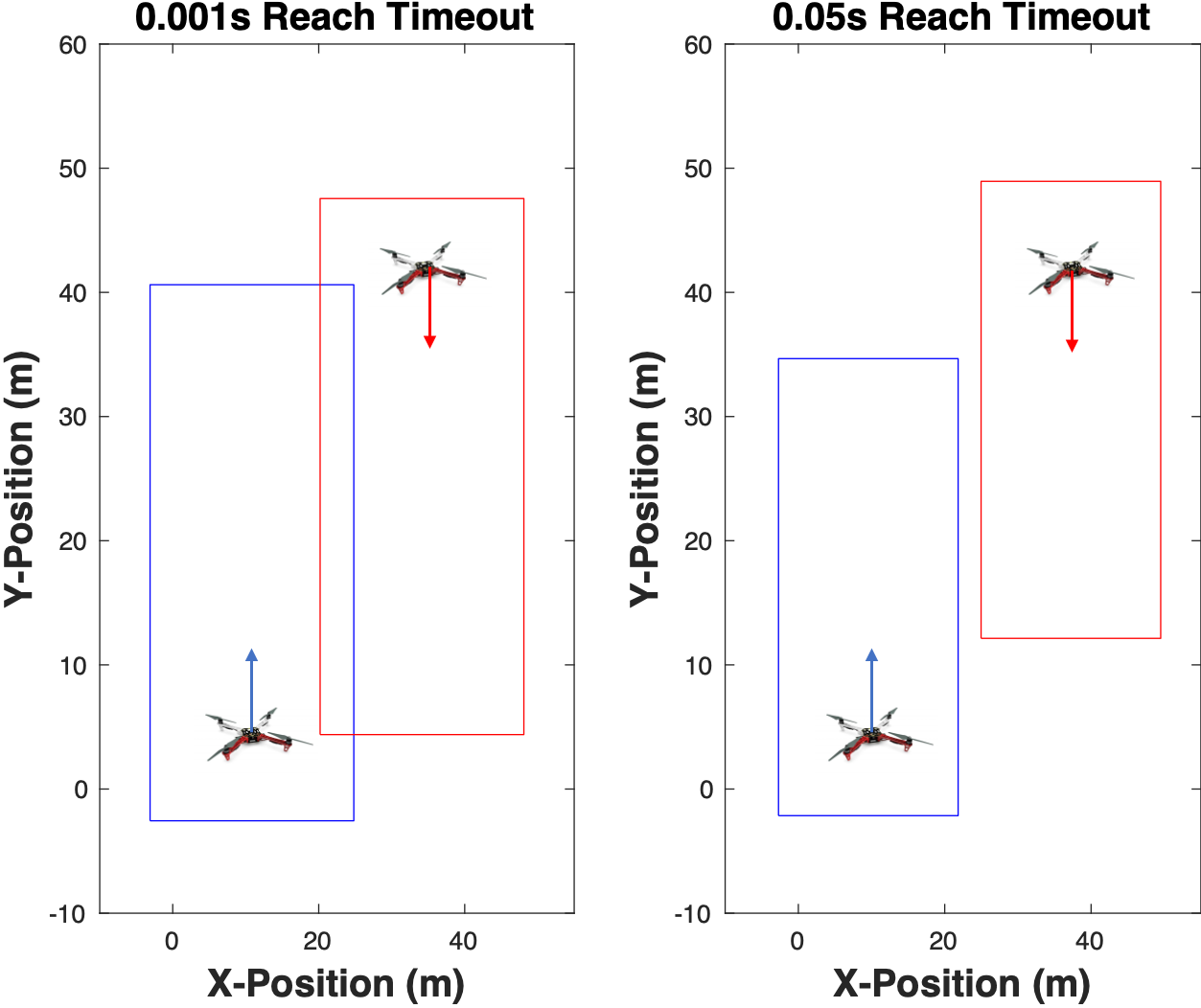}
    \caption{Reachable sets for two agents computed using a \unit[0.001]{s} timeout (left) and a \unit[0.05]{s} timeout (right). The red and blue reachable sets were computed using the same respective agent states, reach time, and initial step size, however, the resultant reachable sets using a \unit[0.001]{s} timeout indicate a potential collision, whereas the \unit[0.05]{s} reachable sets do not indicate a potential collision.}
    \label{fig:iterative_improvement}
\end{figure}

 Algorithm \ref{alg:rs_comp} outlines the computation of a single reachable set for a swarm agent, $A_i$. Let $I_i$ and $t^i$ be agent $A_i$'s initial set and current clock time, respectively, and let $T$, $h_i$, $h_{min}$, and $T_{runtime}$ be runtime parameters that specify the lookahead time, initial step size, minimum step size, and maximum runtime for reach set computation, respectively. First, the reach time period $\left[t^i, t^i + T\right]$ is divided into steps of size $h_i$. Next, the SFL operation is called on each time step using the swarm agent motion model until the reachable set for the entire period is completely constructed. In this work, a simplified 1$^{st}$-order model,

\begin{equation}
    \label{eq:new_model}
     x_i\left(t^i + h\right) = x_i\left(t^i\right) + \dot{x_i}\left(t^i\right)h
\end{equation}
is used for state prediction where $h$ is a time step and $a_x, a_y$ and $a_z$ are acceleration measurements taken from the vehicle's inertial measurement unit. The remaining time allowed for reach set improvement is updated, if \textit{enough} time remains for the next iteration to complete, the step size is halved and the face-lifting method is called iteratively until further iterations cannot be computed to completion or the step size reaches the minimum $h_{min}$. The reachable set accuracy improvement between different reach timeouts can be seen in Figure \ref{fig:iterative_improvement}. Immediately after computation, the reachable set is encoded into a message and broadcast to the other swarm agents.

\begin{algorithm}

\caption{Reachability analysis for agent $A_i$}
\label{alg:rs_comp}

\begin{flushleft}

\textbf{Input: $I_i, t^i, T,h_i, h_{min}, T^i_{runtime}$} \\
\textbf{Output: $\mathcal{R}_i[t^i,t^i+T]$}

\end{flushleft}

\begin{algorithmic}[1]

\State \textbf{procedure} INITIALIZATION
\State  \quad $step=h_i$ \Comment{Reach time step}
\State \quad $T^i_1 = T^i_{runtime}$ \Comment{Remaining runtime}
\State \textbf{procedure} REACHABILITY ANALYSIS
\While {enoughTime() \&\& $step > h_{min}$}
    \State $\mathcal{CR} = I_i$ \Comment{Current reachable set}
    \State $T^i_2=T$ \Comment{Remaining reach time}
        \While{$T^i_2 > 0$}
            \State $\mathcal{R,T'} = SFL(\mathcal{CR}, step, T^i_2,I_i)$
            \State $\mathcal{CR} = \mathcal{R}$ \Comment{Update reach set}
            \State $T^i_2=T'$ \Comment{Update remaining react time}
            \State $\mathcal{R}_i[t_i,t_i+T] = \mathcal{CR}$
        \EndWhile
        \State $T^i_1=T^i_1-(A_i.currentTime()-t^i)$
        \State $step = step/2$ \Comment{Update remaining reach time}
    \EndWhile
\State \textbf{return} $\mathcal{R}_i[t_i,t_i+T]$
\end{algorithmic}
\end{algorithm}

\subsubsection{Peer-to-Peer Collision Avoidance}

Each verification period consists of a single reachable set computation bounded by $T_{runtime}$, this leaves remaining time within $T_v$ for peer-to-peer collision detection and avoidance. After its computation, a reach set is encoded and broadcast to the rest of the swarm. When a message, $m_j$, that was broadcast from agent $A_j$ is received by an agent, $A_i$, the collision avoidance sub-routine in Algorithm \ref{alg:col_check} is called and will begin executing peer-to-peer collision detection. Once $m_j$ is received, it is decoded into a reachable set, $\mathcal{R}_j[t^j,t^j+T]$. After decoding, the received set, $\mathcal{R}_j[t^j,t^j+T]$, and local set, $\mathcal{R}_i[t^i,t^i+T]$, are tested for both usefulness and potential collisions. Usefulness tests are required since time delay during message transmission is non-deterministic and there is no guarantee the message will be delivered before the reach-time horizon, $t^j+T$. If a message is not received before its reach time, at best, we conclude that an agent's safety is uncertain. Agent-to-agent local clock mismatch is mitigated by synchronizing the local clocks to the coordinated universal time (UTC) by the GPS module during initialization. We assume that all local clocks are synchronized to UTC, and the set usefulness test from agent $A_i$'s perspective reduces to:
\begin{equation}
    \label{eq:cond1}
    t_i < t_j + T.
\end{equation}
If the received set satisfies Equation \eqref{eq:cond1}, the algorithm continues with peer-to-peer collision detection.

Potential collisions between a sender and receiver are detected by testing their reachable sets for intersections with each other,
\begin{equation}
    \label{eq:cond2}
    \mathcal{R}_i \cap \mathcal{R}_j = \emptyset.
\end{equation}
If the received reachable set satisfies Equation \eqref{eq:cond2}, the receiving agent, $A_i$, is safe from potential collisions until $t^j + T$ with respect to the sender $A_j$. If Equation \eqref{eq:cond2} is not satisfied, a potential collision has been detected between $A_i$ and $A_j$ and collision avoidance action should be taken.

\begin{algorithm}
\caption{Peer-to-peer collision verification for agent $A_i$}
\label{alg:col_check}

\begin{flushleft}
\textbf{Input: $\mathcal{R}_i[t^i,t^i+T]$, $m_j$, $t_i$} \\
\textbf{Output: $collision,$ $useful$}
\end{flushleft}
\begin{algorithmic}[1]

\State \textbf{procedure} INITIALIZATION
    \State  \quad $collision=false$ \Comment{Reach time step}
    \State \quad $useful=false$ \Comment{Remaining runtime}
\State \textbf{procedure} REACHABILITY ANALYSIS

\If {new message $m_j$ arrive}
    \State $\mathcal{R}_j[t^j,t^j+T]$ = decode($m_j$)
    \State $t^i_{rs} = \mathcal{R}_i.t^i_{rs}$
    \State $t^j_{rs} = \mathcal{R}_j.t^j_{rs}$
    \If{$t^i_c < t^j_{rs} + T$}
        \State \textit{useful} = true
        \State \textit{collision} = checkIntersections($\mathcal{R}_i,\mathcal{R}_j)$
    \EndIf
\EndIf
\State \textbf{return} \textit{collision, useful}
\end{algorithmic}
\end{algorithm}

\section{Experimental Setup and Evaluation} \label{section:experimental_setup_and_evaluation}

In this section, we outline the software implementation of Algorithm \ref{alg:rs_comp} and Algorithm \ref{alg:col_check}, and how simulation was used to facilitate development of our application and prepare it for field deployment. Additionally, the full-scale experimental setup with our UAV platform from section \ref{subsec:system_overviewd} is detailed.

\subsection{Collision Avoidance Application}

Algorithms \ref{alg:rs_comp} and \ref{alg:col_check} were implemented as a C++ library, \texttt{safety\_checker} (link to github furnished upon acceptance), and used to build a runtime verification application within ROS framework. The application consists of several processes (as ROS nodes) connected by message channels in a publisher/subscriber paradigm. Each node's functionality is briefly described here.

\begin{table}[h]
\centering
\caption{Reachable Set Computation Parameters}
\begin{tabular}{@{}l l@{}}
\toprule
\textbf{Parameter} & \textbf{Explanation of Parameter} \\
 \midrule
 Position Error $x_e$ & Minimum  dimensions $\mathcal{R}_i$ \\
 \midrule
 Velocity Error $v_e$ & Minimum time growth rate for $\mathcal{R}_i$\\
 \midrule
 Initial time step size $h_i$ & Initial face-lifting resolution \\
 \midrule
 Reach Time $T$ & Time to forwardly compute $\mathcal{R}_i$\\
 \midrule
 Timeout $T_{runtime}^i$ & Time to iteratively recompute $\mathcal{R}_i$\\
 \bottomrule
\end{tabular}
\label{table:parameters}
\end{table}

\subsubsection*{MAVROS} The software bridge between the ROS workspace and the FCU. Through MAVROS, the vehicle's sensor data is sent into the ROS workspace, making it available to be read by other nodes in the application. Additionally, flight commands such as \texttt{BRAKE}, \texttt{LAND}, \texttt{RETURN}, and \texttt{ALL\_CLEAR} can be sent from other nodes to MAVROS, which forwards them to the FCU to be executed.

\subsubsection*{POS\_XFORM} Transforms the vehicle's GPS coordinates from latitude, longitude, and altitude to the rectangular frame utilized during the reachable set calculations.

\subsubsection*{VERIFIER} Periodically computes the host agent's reach set using the face-lifting method. The characteristics of the reach set can be tuned by specifying computation parameters at runtime; these parameters are listed in Table \ref{table:parameters}.  Additionally, VERIFIER also takes in decoded messages and performs peer-to-peer collision checking as they arrive.

\subsubsection*{SWARM\_COMMS} The software bridge between the ROS workspace and the XBee radio transceiver. The host agent's reach sets are serialized to bytes and broadcast to the other agents, while received reach sets are de-serialized and sent to the VERIFIER node.

\subsection{Containerization}

The ROS application is packaged and distributed to agents as a Docker\footnote{\url{https://www.docker.com/}} container. Software containerization has proven particularly powerful when working with UAV swarms as it guarantees that each system will run the application in identical virtual environments, regardless of the host's architecture, allowing the application to be readily, and rapidly deployed between different host systems. Furthermore, a containerized application can be tested rigorously offline, in simulation, prior to deployment on hardware and conducting costly field testing and experimentation. A container image that passes simulation testing can be pushed directly to the hardware platform with confidence in the application's functionality. This approach to development and testing can greatly speed up software deployment for real, decentralized, robotic systems.

\subsection{Simulation}

We evaluated the decentralized collision avoidance application in both simulation and on a quadrotor swarm at a representative outdoor testing site. For preliminary evaluation, simulation was used to facilitate development and tune the reach set computation parameters shown in Table \ref{table:parameters}. Simulation is an invaluable tool when working with real, complex distributed systems because it provides an alternative testing mechanism to the costly (in time and money) process of organizing field experiments during the software development stage, and allows us to more easily rule out hardware uncertainty while debugging \cite{arnold2019what}.

We have developed a custom simulator targeting aerial swarm deployments that leverages our containerization strategy. Individual vehicles are simulated with ArduPilot's\footnote{\url{http://www.ardupilot.org/}} software-in-the-loop (SITL) simulator. Each simulated vehicle is networked to a Docker container running the anti-collision application -- the same container that will be deployed on the actual vehicle. The simulated vehicles are also networked via a MAVLINK router that serves as the inter-agent communication hub,  which does \textit{not} reflect the transmission delay and dropout that the actual inter-agent communication network may experience, but rather simulates the ideal swarm networking scenario.

\begin{figure*}[t]
     \centering
     \begin{subfigure}[t]{0.325\textwidth}
         \centering
         \includegraphics[width=\textwidth]{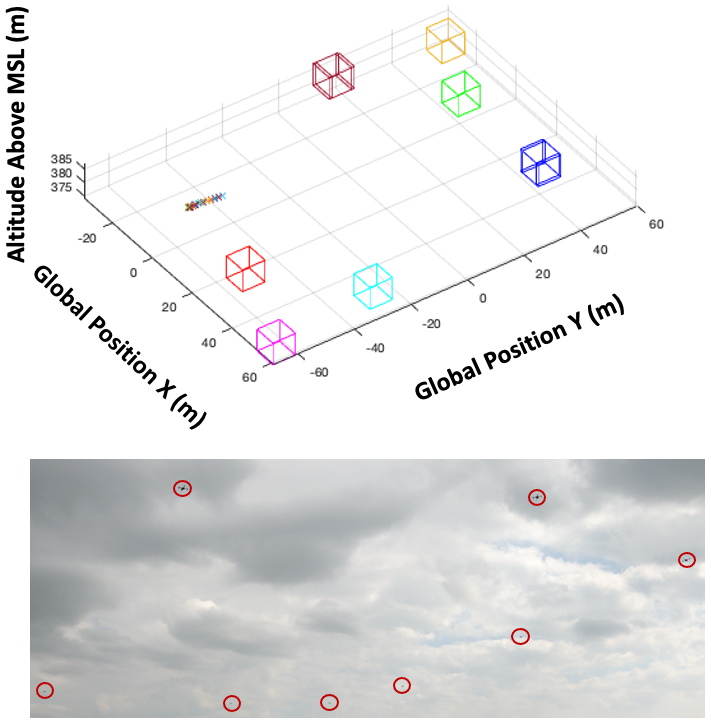}
         \caption{Agents in formation pre-rendezvous.}
         \label{fig:test_left_col}
     \end{subfigure}
     \hfill
     \begin{subfigure}[t]{0.325\textwidth}
         \centering
         \includegraphics[width=\textwidth]{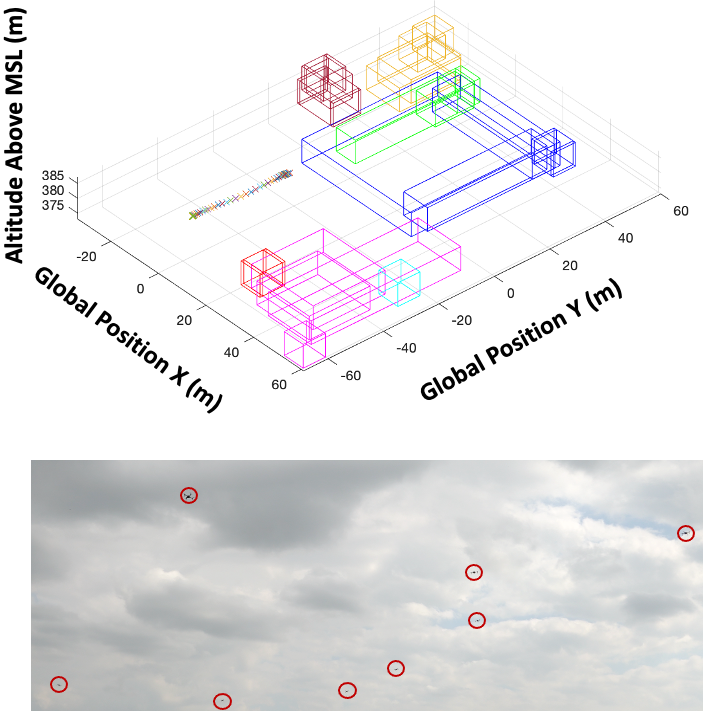}
         \caption{Agents en-route to rendezvous.}
         \label{fig:test_mid_col}
     \end{subfigure}
     \hfill
     \begin{subfigure}[t]{0.325\textwidth}
         \centering
         \includegraphics[width=\textwidth]{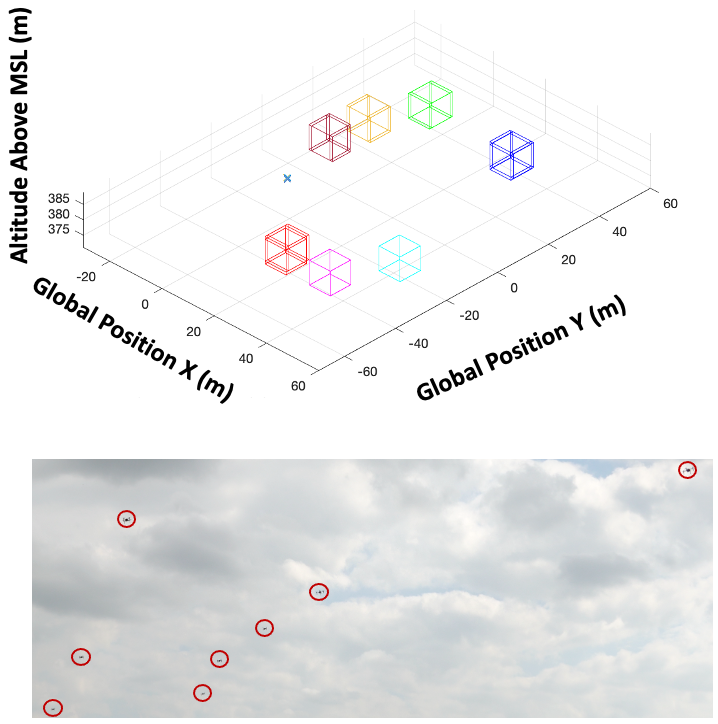}
         \caption{Agents halted after collision detection.}
         \label{fig:test_right_col}
     \end{subfigure}
        \caption{Frames from the video recording of an outdoor collision avoidance experiment are matched with $\sim\unit[10]{s}$ of reachable sets reconstructed from a single agent's logs. (\ref{fig:test_left_col}) Agents executing planned tracking mission, their reachable sets do not intersect and no potential collisions are detected within the reach-time horizon. (\ref{fig:test_mid_col}) the mission is interrupted and all agents are commanded to a rendezvous point which would cause a swarm-wide collision. As the agents move towards each other, the reachable sets begin to take shape and intersect. The intersecting reach sets indicate a collision will occur and avoidance action is taken. (\ref{fig:test_right_col}) After collision avoidance action is taken, the agent's reachable sets shrink back down as they wait for the \texttt{ALL\_CLEAR} from the GCS.}
        \label{fig:outdoor_test}
\end{figure*}

\subsection{Parameter Selection}

The reachable set computation parameters from Table \ref{table:parameters} were initially chosen based on results from a simulated tracking mission and later refined in outdoor experiments.

\subsubsection*{Position and Velocity Error}

The position and velocity error parameters, chosen based on the maximum expected error in GPS measurements, were set to \unit[1.5]{m} and \unitfrac[1.5]{m}{s} respectively.

\subsubsection*{Reach Time}

During the simulated tracking missions it was found that a reach time of \unit[4.0]{s} was sufficient to detect collisions and execute avoidance action with a \unit[0.5]{s} verification period when the agents were moving at their maximum velocity (\unitfrac[10]{m}{s}). For our experiments, the avoidance action consisted of a stop-and-hold maneuver. In future efforts, more complex avoidance action, such as de-conflicted path-planning \cite{herbert2017fastrack}, can be used to prevent collisions with less disruption to the original flight plan.

\subsubsection*{Initial Reach Time Step Size}

The initial time step size was chosen to be the the full length of the reach time (\unit[4.0]{s}); in other words, the first iteration of reachable sets computed before the timeout period has elapsed is the fastest possible computation and the largest over-approximation given the reach time.

\subsubsection*{Timeout}

The timeout parameter was chosen by considering the diminishing improvements between subsequent face-lifting operations. We expect the estimated reachable set size to ``shrink'' between iterations, however, at some point the change in size between iterations becomes insignificant to the time needed to compute them. In order to find the timeout that gives a reasonably conservative reach set in the shortest amount of time, the face-lifting iterative improvement algorithm was run using vehicle states that would produce the largest reachable sets for our chosen \unit[4.0]{s} reach time (i.e., maximum velocity and acceleration) and the change in reachable set dimension between iterations was tracked. It was observed that the reachable set stops improving (shrinking) within the \unit[1.5]{m} GPS error threshold around \unit[0.004]{s} of runtime. With this observation, we have chosen a slightly larger \unit[0.005]{s} reach timeout.

\subsection{Field Deployment}

Field experiments took place in two phases. Each phase of experimentation sought to evaluate 1) the algorithm's performance on fielded hardware, and 2) the application's ability to prevent collisions in a representative, uncertain outdoor environment. The first phase was used to validate the initial verification parameters and gather application runtime data. Timing analysis from the first phase of field experiments informed strategies to improve the application's performance. These improvements were deployed in a second phase of experiments. The algorithm's performance was evaluated by measuring the time for each sub-task listed in Table \ref{table:performance_measures} that compose the verification application.
\begin{table}[h]
\centering
\caption{Anti-Collision Algorithm Timing Performance}
\begin{tabular}{@{}l l@{}}
\toprule
\textbf{Measurement} & \textbf{Expression} \\
 \midrule
 Reachable set computation time & $\tau_{rs}^i = t_e^i - t^i_{rs}$ \\
 \midrule
 Encoding Time & $\tau_{e}^i = t_{tf}^i - t^i_{e}$ \\
 \midrule
 Broadcast time & $\tau_{b}^i = t_{d}^j - t^i_{tf}$ \\
 \midrule
 Decode time & $\tau_{d}^i = t_{c}^i - t^i_{d}$\\
 \midrule
 Collision time & $\tau_{c}^i = t_{f}^i - t^i_{c}$\\
 \bottomrule
\end{tabular}
\label{table:performance_measures}
\end{table}

The collision avoidance application was tested in flight by programming the UAVs to execute a simple ground-target tracking mission, covering a total area of $\unit[7800]{m^2}$. The mission, as planned, does not involve any maneuvers that would be flagged as a potential collision, and the agents are each commanded to the same altitude at the start of the mission. With the collision avoidance application running, a signal is sent from the ground control station (GCS) that immediately commands the vehicles to a rendezvous point (e.g., same latitude, longitude, and altitude), representing what would be an erroneous command in the swarm controller and result in a collision. When an individual vehicle detects a potential collision, the FCU switches flight modes to immediately \texttt{BRAKE} until an \texttt{ALL\_CLEAR} is sent from the GCS, and the original tracking mission resumes.

Figure \ref{fig:outdoor_test} shows an example of a flight test in three panels representing, from left to right respectively, before, during, and after a forced collision is attempted. Approximately \unit[10]{s} worth of received reachable sets from a single agent's perspective are reconstructed from the logs and matched with a frame from a ground-camera recording of the same test. (Figure \ref{fig:test_left_col}) Prior to the rendezvous, the vehicles are moving slowly in a tracking formation as part of the mission; the reachable sets are nearly the size of the initial set $I$. (Figure \ref{fig:test_mid_col}) After GCS commands \textit{all} vehicles to local coordinates (X: 0, Y: 0, alt: 380), the sets expand as the vehicles' velocities increase, and the reachable sets begin to intersect, triggering the avoidance action. (Figure \ref{fig:test_right_col}) The avoidance action successfully switches the vehicles to \texttt{BRAKE}, and their reachable sets shrink back to the minimum set as they hover in place and await the \texttt{ALL\_CLEAR} from the GCS.

\section{Results} \label{section:results}

The verification timing data collected from both phases of flight experiments is used to analyze the impacts of SWaP-constrained hardware on the application as well as inform design recommendations for other verification-based safety applications. During flight experiments, each agent locally measured and logged the verification task times described in Table \ref{table:performance_measures} along with any reachable sets it sent and received. At the conclusion of the field experiments, the logs were pulled from each agent and the following analysis of the broadcast delay and application performance was conducted offline. The remainder of this section is dedicated to analyzing the timing of critical verification tasks from outdoor experiments, and using the analysis to improve the performance of the application.

\subsection{Broadcast Delay}

\vspace{0.5in}
\begin{figure*}[h]
    \centering
    \includegraphics[width=0.85\textwidth]{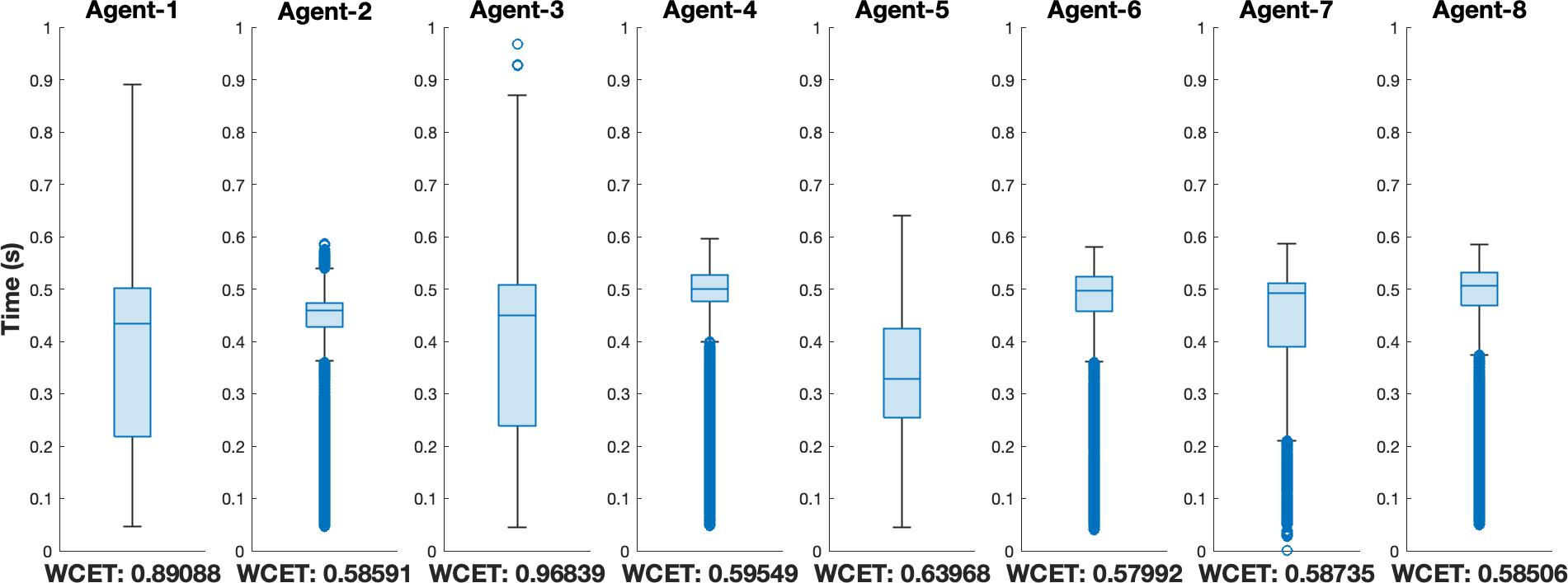}
    \caption{Broadcast time logs and error bars for the 8 agents used in the first phase experiment. Interestingly, despite identical hardware, software, and configuration, Agent-1 and Agent-3 have significantly different broadcast times than other agents with a lower average broadcast time, but a much larger worst case broadcast time.}
    \label{fig:broadcast_times}
\end{figure*}

The transmission and reception of reachable sets is a crucial component of this approach to safety verification. Lightweight, low-cost, low-power radios have limited range, bandwidth, and less-robust communication protocols but enable decentralized communication and mesh networking. However, increased transmission delay and packet loss are problematic, and the extent to which messages are lost or delayed is difficult to predict without real-world data. Although a solution may be to utilize more-capable hardware, alternative communications from Rajant or Doodle Labs are relatively power-hungry and would further burden our SWaP-constrained system to the effect of severely reduced flight times.

The broadcast times (defined in Table \ref{table:performance_measures}) for each agent were recorded for messages received during the flight experiments. Figure \ref{fig:broadcast_times} visualizes the distribution of broadcast times recorded during an 8-vehicle flight experiment. For this flight test, all messages were received within \unit[1]{s}, and the average broadcast times are between \unit[0.3-0.6]{s}. This timing information is critical to designing a sound verification application, specifically the choice in reach-time horizon; if the reach-time horizon is set too low, the transmission delay may result in a majority of sent reachable sets expiring before they are received, and no meaningful collision detection can take place.

Furthermore, we can see that the average broadcast times are consistent from agent to agent, however, the time range for Agent-1 and Agent-3 are noticeably larger than the six other agents' ranges. While deeper investigation into the cause of this difference is beyond the scope of this work, it further highlights that the non-deterministic nature of less-capable communication systems needs to be taken into account at design time. It also highlights how identically designed, and implemented hardware and software can still result in variation, and that discrepancies in even one agent's component can impact timing in a robotic swarm. Further discussion related to mitigating message loss due to poor network performance can be found in Section \ref{sec:mitigating-lost-messages}.

\subsection{Worst-Case Execution Time Insights}

When designing safety-critical runtime systems, it is paramount to consider the worst-case execution time (WCET) for each task to ensure the system will still function in the worst-case scenario. The WCET for each verification sub-task across all agents is reported in Table \ref{table:p1_test_results} for the first phase and Table \ref{table:p2_test_results} for the second phase. Overall, the WCET for each task does not vary significantly from agent to agent and it can be seen that the reach set computation requires the most (processor) time to complete.

\begin{table*}[h]
\centering
\caption{Second Phase: Worst-Case Execution Times for Runtime Verification Sub-Tasks}
\begin{tabular}{ @{ }l l@{ } l@{ } l@{ } l@{ } l@{ } l@{ } l@{ } l@{ } }
\toprule
\textbf{Sub-Task} & \textbf{Agent 1} & \textbf{Agent 2} & \textbf{Agent 3} & \textbf{Agent 4} & \textbf{Agent 5} & \textbf{Agent 6} & \textbf{Agent 7} & \textbf{Agent 8}\\
 \midrule
 Reach Set Time ($\tau_{rs}$) (s) & 5.1e-3 & 5.2e-3 & 5.2e-3 & 5.3e-3 & 4.7e-3 & 5.3e-3 & 4.9e-3 & 5.2e-3\\
 \midrule
 Encoding Time ($\tau_{e}$) (s) & 7.2e-06 & 7.2e-06 & 6.7e-06 & 7.4e-06 & 7.6e-06 & 7.4e-06 & 6.9e-06 & 7.2e-06\\
 \midrule
 Broadcast Time ($\tau_{b}$) (s) & 2.205 & 2.529 & 2.772 & 2.112 & 2.223 & 2.152 & 2.384 & 1.923\\
 \midrule
 Decode Time ($\tau_{d}$) (s) & 4.1e-05 & 4.6e-05 & 3.1e-05 & 2.6e-05 &  7.2e-05 & 6.5e-05 &  5.3e-05 & 5.3e-05\\
 \midrule
 Collision Time ($\tau_{c}$) (s) & 6.4e-04 & 6.5e-04 & 6.0e-04 & 6.4e-04 & 6.5e-04 & 5.8e-04 & 7.1e-04 & 6.8e-04\\
 \bottomrule
\end{tabular}
\label{table:p2_test_results}
\end{table*}

We have identified design parameters that affect the applications performance, namely the number of distributed agents ($N$), the reach set timeout ($T_{runtime}$), reach time ($T$), and the verification period ($T_v$). While the choice of parameter values are specific to the needs of a particular deployment, we can identify the bounds and characterize the effects of the different parameter values.

\subsubsection{Bounds on the Number of agents}

First we determine the bounds on the number of agents ($N$) the application can accommodate. Intuitively, the lower bound is $N=2$. To determine the upper bound, we must consider the entire verification period, $T_v$, which consists of all sub-tasks in Figure \ref{fig:reach_timeline}. The sub-tasks can be categorized as either time allotted for reachable space calculation and message encoding, $T_{rs}$, or to message decoding and collision checking, $T_c$, where:
\begin{equation}
\label{eq:reachable_space_time}
    T_{rs} = \tau_{rs} + \tau_{e}
\end{equation}
\begin{equation}
\label{eq:collision_check_time}
    T_c = \tau_{d} + \tau_c.
\end{equation}

The verification period, $T_v$, consists of one occurrence of $T_{rs}$ and $N-1$ occurrences of $T_c$. Assuming the entire verification period is completely utilized for verification tasks, we can state that the time dedicated to reachable state calculation and encoding in addition to the time spent decoding and collision checking $N-1$ agents must be no greater than the verification time period, or

\begin{equation}
\label{eq:verificaion_time}
    T_v > T_{rs} + (N-1)T_c
\end{equation}

\noindent By substituting Equations \eqref{eq:reachable_space_time} and \eqref{eq:collision_check_time} into Equation \eqref{eq:verificaion_time} and solving for $N$ we get an expression for the theoretical upper bound:

\begin{equation}
\label{eq:solved_for_N}
  N \leq \frac{T_v - \tau_{rs} - \tau_e}{\tau_d + \tau_c} + 1.
\end{equation}

We construct an estimate for the maximum number of agents, $N_{max}$, the verification loop can handle by choosing the largest sub-task execution times from Table \ref{table:p1_test_results} across all agents and substitute them into Equation \eqref{eq:solved_for_N} along with our chosen verification period $T_v=\unit[0.5]{s}$.

Based strictly on the computational capacity of the companion computer, Equation \eqref{eq:solved_for_N} yields a theoretical upper bound of $N_{max}=$ 3968 agents. However, we emphasize that this is a computation-bound limit. In a real-world system, the true bottleneck shifts from the CPU to the communication network well before this limit is reached. As noted in our broadcast delay analysis \ref{fig:broadcast_times}, packet collisions and bandwidth saturation on standard mesh radios (like ZigBee or Wi-Fi) will degrade safety performance significantly earlier. Therefore, while the verification framework is highly scalable computationally, the system scalability is constrained by the network technology. Future work involving swarms of this magnitude would require TDMA-based communication protocols to support the broadcast frequency required for safety.

 \begin{table*}[h!]
\centering
\caption{First Phase: Worst-Case Execution Times for Runtime Verification Sub-Tasks }
\begin{tabular}{ @{ }l l@{ } l@{ } l@{ } l@{ } l@{ } l@{ } l@{ } l@{ } }
\toprule
\textbf{Sub-Task} & \textbf{Agent 1} & \textbf{Agent 2} & \textbf{Agent 3} & \textbf{Agent 4} & \textbf{Agent 5} & \textbf{Agent 6} & \textbf{Agent 7} & \textbf{Agent 8}\\
 \midrule
 Reach Set Time ($\tau_{rs}$) (s) & 1.6e-2 & 1.5e-2 & 1.5e-2 & 1.6e-2 & 1.5e-2 & 1.5e-2 & 1.5e-2 & 1.5e-2 \\
 \midrule
 Encoding Time ($\tau_{e}$) (s) & 1.6e-5 & 1.1e-5 & 1.6e-5 & 1.3e-5 & 9.0e-6 & 1.3e-5 & 1.3e-5 & 1.2e-5\\
 \midrule
 Broadcast Time ($\tau_{b}$) (s) & 0.89 & 0.59 & 0.97 & 0.60 & 0.64 & 0.58 & 0.59 & 0.59\\
 \midrule
 Decode Time ($\tau_{d}$) (s) & 1.2e-4 & 1.2e-4 & 1.2e-4 & 1.3e-4 & 1.2e-4 & 1.3e-4 & 1.3e-4 & 1.3e-4 \\
 \midrule
 Collision Time ($\tau_{c}$) (s) & 1.2e-5 & 7.0e-6 & 1.2e-5 & 7.0e-6 & 7.0e-6 & 7.0e-6 & 7.0e-6 & 7.0e-6\\
 \bottomrule
\end{tabular}
\label{table:p1_test_results}
\end{table*}

\subsubsection{Bounds on the Reachable Set Computation Run Time}

In the previous section, we found that the model for the number-of-agents upper bound depends on the time used to calculate the reachable set, $\tau_{rs}$. Recall that this time is already bounded at design time to a maximum specified runtime, $T_{runtime}=\unit[0.005]{s}$ for our deployment. However, Table \ref{table:p1_test_results} reports the measured WCET for $\tau_{rs}$ much larger than the specified $T_{runtime}$.

A closer look at the face-lifting iterative improvement algorithm from \cite{tran2019decentralized} reveals that a new iteration will be started if there is \emph{any} time left for additional computation, and if so, then the number of time steps is doubled and another face-lifting iteration is run to completion. By only checking if there is any time remaining, depending on the chosen $T_{runtime}$, we risk the scenario where the time used for reachable set computation time exceeds the allotted run time , or worse, the reachable set computation time exceeds the entire verification period, leaving no portion of $T_v$ available for peer-to-peer collision checking.

To prevent these scenarios, we propose using a regression model to predict the time needed to execute subsequent face-lifting iterations and use this model to check \emph{if} there is enough time to complete the next FL iteration (Algorithm \ref{alg:rs_comp}, line 5). The regression model is built by by observing the amount of time needed to execute consecutive face-lifting iterations. Between each face-lifting iteration, the number of steps doubles, or grows exponentially, and as such, we choose to fit a curve of the form $t_{rs}=ae^{bn}$ where $t_{rs}$ is the total reach set computation run time, $n$ is the number of iterations, and $a$ and $b$ are constants. Run-times the first 10 iterations of reachable set computations on a Raspberry Pi were collected, and it was found that the number microseconds needed to compute $n$ face-lifting iterations can be modeled by $t_{rs} = 416.8457e^{0.6537n}$. Due to the non-deterministic nature of process scheduling, subsequent compute times will likely not match the data exactly. With this in mind, we add a 5\% buffer to the predictive model to slightly over-approximate the compute time.

We can now construct bounds for reach set computation time, $T_{rs}$, using the regression model. The lower bound can be set to 0 as it will always complete at least one iteration, regardless of the time given and the upper bound can be set to our regression model with a $5\%$ buffer, we can write the bounds on $T_{runtime}$ as:
\begin{equation}
    \label{eq:exp_regression}
    0 \leq T_{runtime} \leq 437.6879e^{0.6537n}
\end{equation}
\

Equation \eqref{eq:exp_regression} was used in the second phase of experiments and the improved reach set computation times reported in Table \ref{table:p2_test_results}. With respect to timing performance, the reach set computation time is the primary improvement, as the WCET was reduced from surpassing the targeted \unit[0.005]{s} by \unit[0.010]{s} to being within \unit[0.001]{s} of the target time.

\subsubsection{Mitigating Lost Messages (selecting $T$ and $T_v$)}
\label{sec:mitigating-lost-messages}

During the broadcast time analysis, we found that, on average, each agent was only receiving about 60\% of the expected reachable state messages. This packet loss rate is dangerously high for any safety-critical system, and given the non-deterministic nature of our mesh network, there is no way to guarantee message delivery between agents. We may however, attempt to design the verification loop to mitigate the risk of a message not being delivered.

Recall that the encoded reachable set message is sent once during each control period, effectively making $T_v$ the rate the messages are sent, and let $A_i$ and $A_j$ be two networked agents that calculate sets $\mathcal{R}_i[t^i_1,t^i_1+T]$ and $\mathcal{R}_j[t^j_1,t^j_1+T]$ respectively. Furthermore, assume that $A_i$ is stationary and $A_j$ is moving rapidly towards $A_i$ and they \textit{will} collide if no action is taken. The time to collision (TTC) from $A_j$'s perspective can be expressed as:
\begin{equation}
    \label{eq:ttc}
    TTC_1^j = (t^i_1 + T) - t^j_1.
\end{equation}
If $A_j$ does not receive the reach set message from $A_i$, no potential collision is detected at this time, and if no action is taken, the collision will still occur. We know that the next round of reachable sets will be calculated and broadcast in one verification period, $T_v$. If we assume that the next set, and all subsequent reach sets sent from $A_i$ will indicate a collision, we can say that TTC will decrease by $n-1$ verification periods until the collision is detected or TTC reaches zero:
\begin{equation}
    \label{eq:ttc_tv}
    TTC_1^j - (n-1)T_v = (t^i_1 + T) - t^j_1 - (n-1)T_v.
\end{equation}
Equation \eqref{eq:ttc_tv} can be used to derive a relationship between the verification period and the reach time parameters that can be used to reduce the risk of a collision occurring due to undelivered messages. Recall that there is a 40\% chance a message will not be delivered, which means that if 5 messages messages are sent in a row, there is a $1-0.4^{5}=99\%$ chance at least one of the messages will be received. Therefore, if a collision \textit{will} occur within some time window TTC, there is a 99\% chance the agents at risk can communicate if 5 messages are sent before TTC reduces to zero. If we assume that after $n=5$ iterations the left-hand side of Equation \eqref{eq:ttc_tv} goes to zero, we are left with the relationship:
\begin{equation}
    \label{eq:ttc_0}
    0 = t^i_1 + T - t^j_1 - 4T_v.
\end{equation}
Recall that $t^i$ is the time the reachable set is computed on $A_i$, and $t_j$ is the time the message is received, which simplifies to the broadcast time $\tau_b$\footnote{Decode and encode times are ignored here as $\tau_d$ and $\tau_e$ are 5 orders of magnitude less than the broadcast time} defined in Table \ref{table:performance_measures}. Then,
\begin{equation}
    \label{eq:tvt}
    T_v = \frac{\tau_b + T}{4}.
\end{equation}
We can say that if the choice of $T_v$ and $T$ satisfies the relationship in Equation \eqref{eq:tvt}, there is a 99\% chance a potential collision will be detected given the message loss rate of our network.

\subsection{Collision Avoidance}

During experimentation, all 140 deliberate attempts to force a collision were detected, on average, \unit[3.2]{s} before they would have occurred. Our in-depth timing analysis of critical verification tasks on an imperfect CPS platform coupled with 100\% of collisions avoided indicate that this is a sound method for assured swarm safety.

\section{Conclusions} \label{section:conclusions}

This work has highlighted the importance of considering the limitations of electronic systems when attempting to make safety guarantees and how a theoretically sound solution can fall short of meaningful results if these limitations are neglected. The effectiveness of the swarm-safety application presented in this work, whose core algorithmic and theoretical principles have been previously established and validated in solely simulations, was found to be highly dependent on the capability of the target platform and required data from extensive real-world experimentation before full-scale deployments could be safely conducted. The primary contribution of this work lies not in algorithmic novelty, but in the system-level synthesis and rigorous experimental characterization required to bridge the gap between formal theory and outdoor deployment.

Through field experimentation, we demonstrated that decentralized formal verification is viable for real-time safety in robotic swarms. Crucially, our analysis reveals that the bottleneck for scalability is not computational complexity, as often assumed in simulation, but rather the stochastic nature of network reliability and broadcast delays. Consequently, future efforts to scale assured autonomy must move beyond algorithmic optimization to embrace a co-design approach where control, verification, and communication layers are integrated to mitigate the physical realities of imperfect hardware.

\bibliographystyle{IEEEtran}
\bibliography{justin,references}

\begin{IEEEbiography}[{\includegraphics[width=1in,height=1.25in,clip,keepaspectratio]{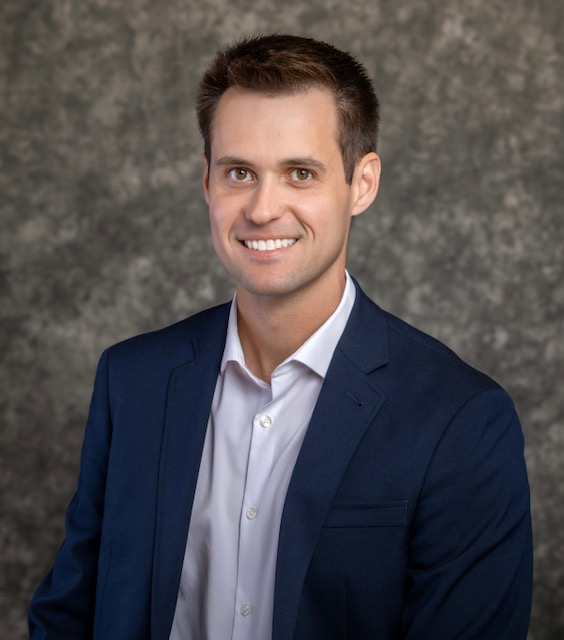}}]%
{Grant Phillips} received his B.S. degree in mechanical engineering from the University of Nebraska-Lincoln, Lincoln, NE, USA, in 2018, and his Ph.D. degree in computer engineering from the University of Nebraska-Lincoln, Lincoln, NE, USA, in 2025. He is currently a Research Engineer with Scientific Systems, Burlington, MA, USA. His research interests include real-time systems, UAS swarm control and coordination, and field robotics.
\end{IEEEbiography}

\begin{IEEEbiography}[{\includegraphics[width=1in,height=1.25in,clip,keepaspectratio]{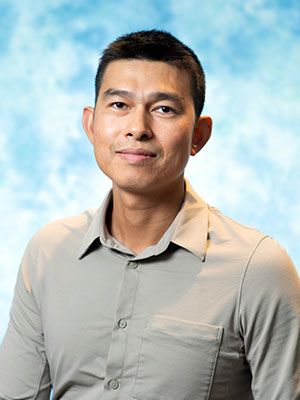}}]%
{Hoang-Dung Tran} earned a Ph.D. degree in Computer Science at Vanderbilt University in August 2020. He is currently an Assistant Professor in the Department of Computer and Information Science and Engineering (CISE) at the University of Florida, Gainesville, FL  USA. His research interests are reachability analysis and robustness certification of deep neural networks (DNN), formal verification of autonomous cyber-physical systems(CPS) with learning-enabled components, safe and robust training for DNN, real-time verification, and safe motion planning for distributed CPS.
\end{IEEEbiography}

\begin{IEEEbiography}[{\includegraphics[width=1in,height=1.25in,clip,keepaspectratio]{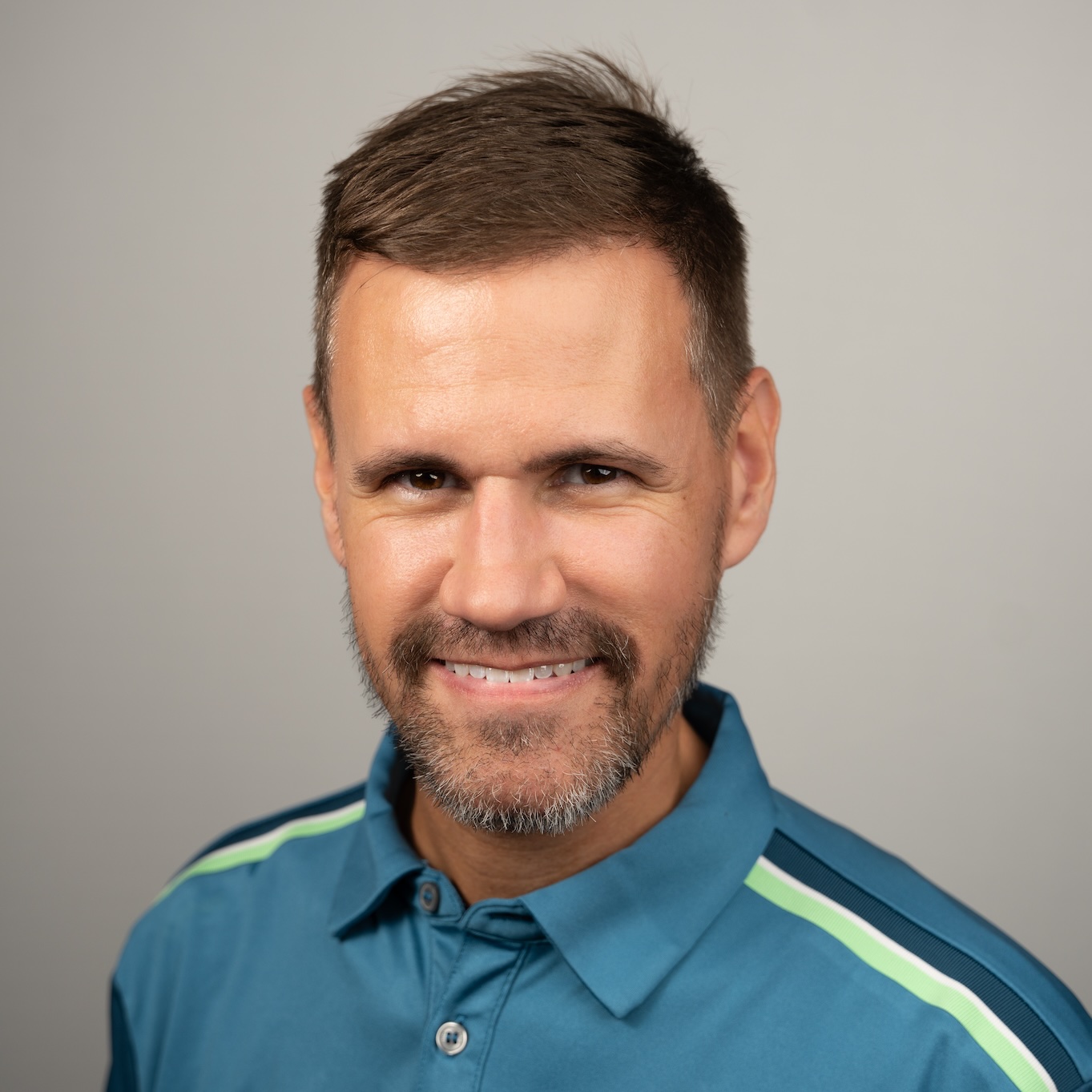}}]%
{Justin M. Bradley} received the B.S. degree in computer engineering from BYU, Provo, UT, USA, in 2005, and the Ph.D. degree in aerospace engineering from the University of Michigan, Ann Arbor, MI, USA, in 2014. He is currently an Associate Professor with the Department of Computer Science, NC State University, Raleigh, NC, USA. His research interests include co-regulation of cyber-physical resources, multi-agent UAS coordination, and CPS safety and assurance.
\end{IEEEbiography}

\end{document}